\documentclass[10pt,twocolumn,letterpaper]{article}

\usepackage{cvpr}              

\usepackage{siunitx}
\usepackage{xcolor}
\usepackage{colortbl}
\usepackage{multirow}
\usepackage{booktabs} 
\usepackage{makecell}
\usepackage{tabularx}
\usepackage{cancel}
\usepackage{amsthm}

\definecolor{cvprblue}{rgb}{0.21,0.49,0.74}
\usepackage[pagebackref,breaklinks,colorlinks,allcolors=cvprblue]{hyperref}
\hypersetup{
    colorlinks=true,
}

\def\confName{CVPR}
\def\confYear{2026}

\title{DiverseGRPO: Mitigating Mode Collapse in Image Generation via Diversity-Aware GRPO}

\author{Henglin Liu$^{1,2*}$,
        Huijuan Huang$^{2\dagger\S}$,
        Jing Wang$^{2,3*}$,
        Chang Liu$^{1}$,
        Xiu Li$^{1\dagger}$,
        Xiangyang Ji$^{1\dagger}$
        \\
        $^{1}$Tsinghua University,
        $^{2}$Kling Team, Kuaishou Technology,
        $^{3}$Shenzhen Campus of Sun Yat-sen University
        \\
        \textit{\href{https://henglin-liu.github.io/DiverseGRPO}{Project Page} \quad
        $^{\S}$Project Leader. \quad
        $^{\dagger}$Corresponding Authors. \quad
        $^{*}$Work Conducted During Internship.} \\
        {\tt\small liu-hl24@mails.tsinghua.edu.cn}
}

\begin{document}

\twocolumn[{
\renewcommand\twocolumn[1][]{#1}%
\maketitle
\vspace{-30pt}
\begin{center}
    \centering
    \includegraphics[width=1.0\linewidth]{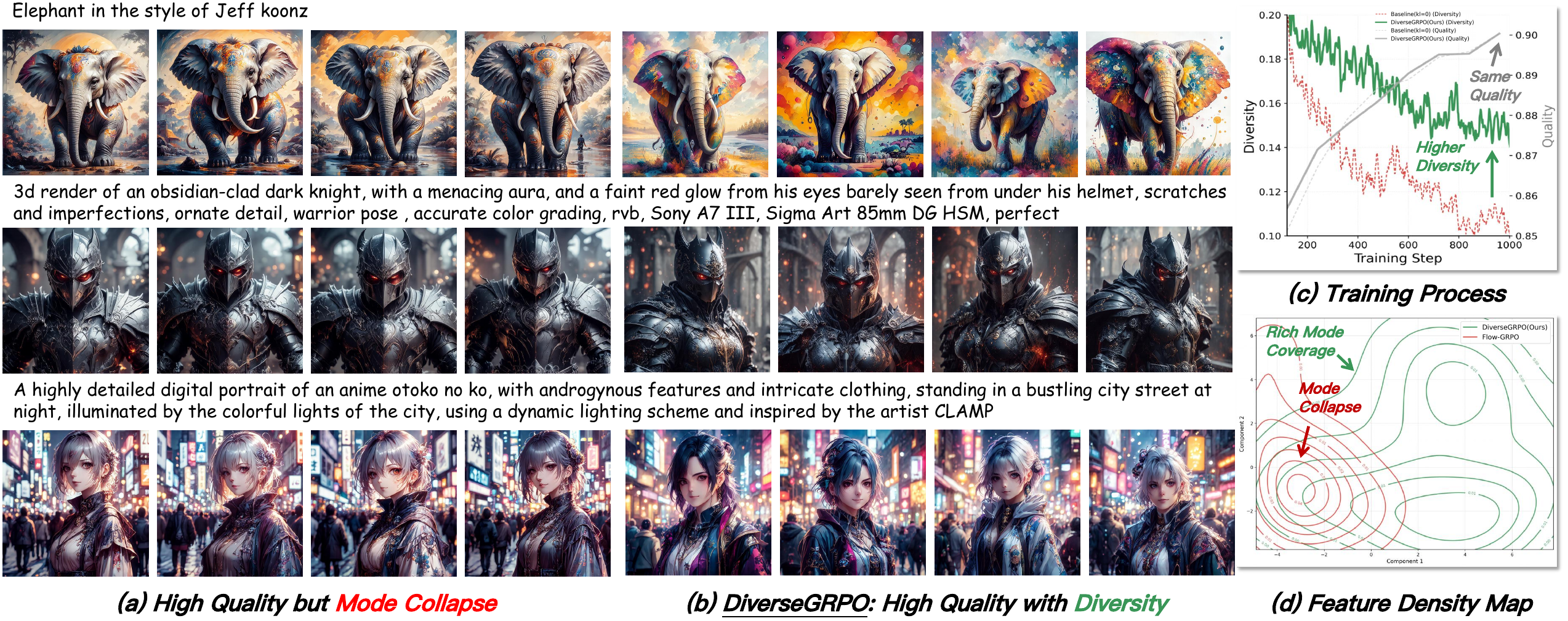}
    \vspace{-5mm}
    \captionof{figure}{(a) Image generation models trained with GRPO suffer from mode collapse (similar faces, camera angles, etc.), which limits their applicability in creative scenarios. (b) The proposed DiverseGRPO method achieves higher diversity while maintaining comparable quality. (c) DiverseGRPO successfully maintains a healthier level of diversity across the entire duration of training, while the baseline method suffers from a premature collapse. (d) In the Inception feature space, DiverseGRPO generates images that cover a significantly broader range of semantic features, effectively mitigating mode collapse.}
    \label{teaser}
\end{center}%
}]
\begin{abstract}

Reinforcement learning (RL), particularly GRPO, improves image generation quality significantly by comparing the relative performance of images generated within the same group. However, in the later stages of training, the model tends to produce homogenized outputs, lacking creativity and visual diversity, restricting the application scenarios of the model.
This issue can be analyzed from both reward modeling and generation dynamics perspectives. First, traditional GRPO relies on single-sample quality as the reward signal, driving the model to converge toward a few high-reward generation modes while neglecting distribution-level diversity. Second, conventional GRPO regularization neglects the dominant role of early-stage denoising in preserving diversity, causing a misaligned regularization budget that limits the achievable quality–diversity trade-off.
Motivated by these insights, we revisit the diversity degradation problem from both reward modeling and generation dynamics. At the reward level, we propose a distributional creativity bonus based on semantic grouping. Specifically, we construct a distribution-level representation via spectral clustering over samples generated from the same caption, and adaptively allocate exploratory rewards according to group sizes to encourage the discovery of novel visual modes. At the generation level, we introduce a structure-aware regularization, which enforces stronger early-stage constraints to preserve diversity without compromising reward optimization efficiency. Experiments demonstrate that our method achieves an 13\%$\sim$18\% improvement in semantic diversity under matched quality scores, establishing a new Pareto frontier between image quality and diversity for GRPO-based image generation.
\end{abstract}
%
\vspace{-15pt} 
\section{Introduction}
\label{sec:intro}

The diversity of generated images is a key criterion for evaluating the performance of generative models. A significant loss of diversity represents a major challenge for the practical application, particularly in creative fields such as digital art, advertising, and game design, where novelty and variety are fundamental to their success.

However, reinforcement learning from human feedback (RLHF)~\cite{bai2022training,casper2023open} techniques for image generation~\cite{black2023training}, such as Flow-GRPO~\cite{liu2025flow} and DanceGRPO~\cite{xue2025dancegrpo}, have achieved remarkable progress in aligning text-to-image generation models with human aesthetic preferences, recent studies~\cite{cui2025entropy,he2025skywork,yu2025dapo} in large language model(LLM) have revealed a critical limitation of GRPO-based approaches: the degradation of generation diversity. As illustrated in Fig.~\ref{teaser}, this phenomenon manifests as homogenized results (nearly identical character appearances and highly similar perspectives), indicating a collapse of semantic diversity. That is because the intrinsic objective of reward maximization tends to overfit the model to a narrow subset of high-reward modes, effectively encouraging the model to reproduce `safe' or `high-score' patterns while suppressing creative or unconventional outputs. 

This observation raises a deeper question: \textbf{Is diversity degradation an inevitable byproduct of reward optimization, or is it a symptom of misaligned learning objectives and generation dynamics?}

\begin{figure}[t]
  \centering
   \includegraphics[width=\linewidth]{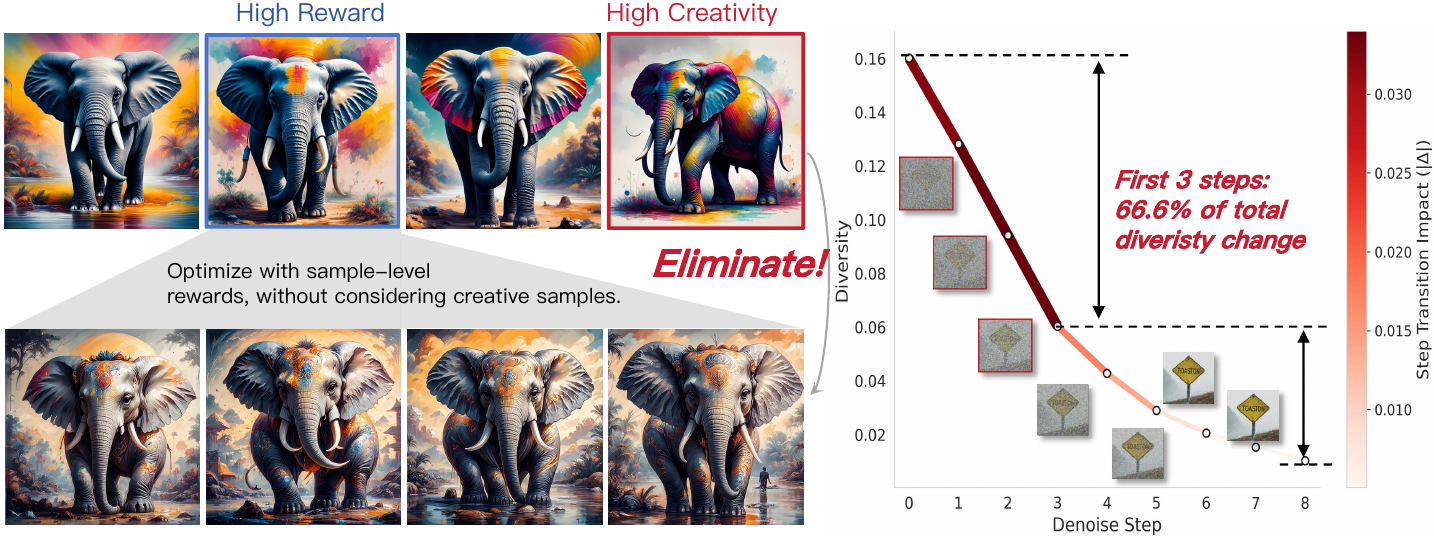}
   \caption{Analysis of the reasons for mode collapse: (Left) Policy model collapse into high-reward modes due to single sample reward modeling. (Right) Conventional regularization neglects the dominant role of early-stage denoising in preserving diversity.}
   \label{motivation}
\vspace{-15pt}   
\end{figure}

We begin by examining the problem from the reward modeling perspective. 
As shown in Fig.~\ref{motivation}.a, GRPO relies solely on single-sample reward signals, where the reward model assigns individual scores to each image without considering the relationships between samples. This approach leads the generative model to become `short-sighted' and `conservative', focusing on maximizing immediate rewards at the expense of exploration and innovation. To analyze this effect more formally, consider that the conditional generation distribution can be decomposed into $K$ semantic modes:$\pi_{\theta}(x \mid p)=\sum_{k=1}^{K} w_{k} \pi_{\theta}^{k}(x \mid p)$, where each component $\pi_{\theta}^{k}$ represents a distinct visual mode (e.g., composition, lighting, or style), and $w_{k}$ is its mixture weight. Let the average reward within each mode be: $ \bar{r}_{k}=\mathbb{E}_{x \sim \pi_{\theta}^{k}}[r(x, p)] $, then the expected reward objective can be rewritten as $ J(\theta)=\sum_{k=1}^{K} w_{k}\bar{r}_{k} $. During optimization, modes with larger average reward $\bar{r}_{k}$ gain higher sampling probability, yielding the following \textbf{replicator dynamics}:$$ \frac{dw_{k}}{dt}=w_{k}\left(\bar{r}_{k}-\mathbb{E}_{j}\left[\bar{r}_{j}\right]\right) $$ This equation describes a self-reinforcing selection process:
modes with slightly higher average rewards continue to grow in weight, while others are gradually eliminated.
At equilibrium, only the dominant high-reward mode survives:
$$
w_{k}=\left\{
\begin{array}{ll}
1, & \text{if } k = \arg\max_{j} \bar{r}_{j}, \\
0, & \text{otherwise.}
\end{array}
\right.
$$
Thus, single-sample reward optimization implicitly drives the model toward a unimodal distribution, leading to homogenized outputs and a collapse in visual diversity. Therefore, to prevent mode extinction, it is insufficient to adjust sampling strategies or model architectures; the reward itself must be redesigned to reshape the underlying dynamics. Instead of rewarding individual samples independently, we introduce a distribution-aware reward that depends on the semantic structure of the generated set for each prompt. Specifically, we construct a distribution-level representation via spectral clustering over samples generated from the same text prompt, and adaptively allocate exploratory rewards according to group sizes, which encourages the discovery of novel visual modes.


In addition to the external reward signal, we further investigate how the intrinsic denoising dynamics of diffusion models influence visual diversity. 
We compute the perceptual similarity between samples generated at different denoising steps, where all samples within the same denoising step share the same latent variable, using the DreamSim~\cite{fu2023dreamsim} model. As shown in Fig.~\ref{motivation}.b, The more denoising steps that are shared, the more similar the samples are, which aligns with previous observations~\cite{ho2020denoising,wang2023diffusion}. However, it is noteworthy that the decline in diversity is steeper during the early denoising phase (as indicated by the red bar, where the first one-third of the denoising steps accounts for approximately 66\% of the overall diversity change). This suggests that the early denoising phase has a disproportionately large impact on the resulting diversity. From the perspective of mitigating mode collapse, the denoising trajectory forms an imbalanced `diversity budget', where early steps are diversity-critical while later steps mainly refine visual quality. Under this perspective, because the diffusion variance is largest in early steps, the KL penalty becomes effectively weakest exactly when the budget should be highest, resulting in a structural mismatch that accelerates mode extinction. To resolve this budget misallocation, we reformulate diversity preservation as a structure-aware regularization scheduling problem, and replace the uniform KL penalty with a Wasserstein constraint that concentrates the regularization budget in the high-impact early phase, while lifting the constraint in the late phase to preserve reward quality and image fidelity. 
To validate the effectiveness and generality of our approach, we conduct extensive experiments across multiple diffusion backbones (SD3.5~\cite{esser2024scaling} and Flux~\cite{blackforestlabs2024flux}) and heterogeneous reward models (PickScore~\cite{kirstain2023pick} and HPSv3~\cite{ma2025hpsv3}). In all settings, our method consistently improves semantic diversity while preserving or even enhancing image quality, outperforming existing GRPO-based baselines under matched reward budgets.

Overall, we conducted an in-depth analysis of mode collapse in GRPO for image generation and innovatively designed a new GRPO training paradigm. Specifically, our main contributions are as follows:
\begin{itemize}
\item \textbf{Diversity-aware reward modeling for mitigating mode collapse.} DiverseGRPO introduces a distribution-level reward formulation that moves beyond conventional single-sample scoring. By applying spectral semantic grouping and assigning exploration bonuses inversely to group size, our approach explicitly incentivizes the discovery and preservation of rare visual modes—directly countering the root cause of mode collapse.

\item \textbf{Structure-aware regularization tailored to diffusion dynamics.} We uncover a fundamental misalignment between standard regularization and the uneven diversity sensitivity along diffusion trajectories. To correct this, we design a Wasserstein-based structure-aware constraint that applies stronger early-step regularization, where diversity is most fragile, while gradually relaxing the penalty in later stages to enhance
the effectiveness of reward optimization.

\item \textbf{State-of-the-art diversity–quality trade-off in GRPO-based image generation.} Comprehensive experiments demonstrate that DiverseGRPO consistently mitigates mode collapse, achieving up to 18\% improvement in semantic diversity under matched quality. Across multiple backbones and reward models, our method establishes a new Pareto frontier for GRPO-driven image generation.
\end{itemize}


\section{Background}
\label{sec:background}

\subsection{RL for Diffusion and Flow Models.}
Building on the success of reinforcement learning (RL) in Large Language Models (LLMs), algorithms such as PPO~\cite{black2023training,schulman2017proximal} and DPO~\cite{wallace2024diffusion} have been adapted to diffusion models for preference alignment and task-specific optimization. This trend has extended to flow-based models. Flow-GRPO~\cite{liu2025flow} and DanceGRPO~\cite{xue2025dancegrpo} integrate GRPO-style policy updates into flow matching, transforming deterministic ODE sampling into stochastic SDE formulations to introduce exploration noise. Subsequent works like MixGRPO~\cite{li2025mixgrpo} propose a hybrid ODE-SDE sampling strategy to improve training efficiency. Addressing the noise inconsistency issue in SDE sampling, Flow-CPS~\cite{wang2025coefficients} introduces a noise-consistent scheme for more accurate reward estimation. Further innovations tackle the challenge of reward sparsity in multi-step trajectories. TempFlowGRPO~\cite{he2025tempflow} move beyond assigning a single global reward. In a parallel and significant development, BranchGRPO~\cite{li2025branchgrpo} 
introduces a tree-structured branching mechanism within the diffusion/flow matching inversion process. It allows multiple trajectories to share prefixes and split at intermediate steps, enabling dense, layer-wise reward fusion.

Most existing methods focus on improving the efficiency of policy optimization but overlook mode collapse issue, which severely diminishes visual diversity and limits the practical applicability of the models. In this work, we conduct an in-depth analysis of this problem and propose an effective solution.
\subsection{Mode-Collapse in Generation Models}
Research on preventing mode collapse in large language models (LLMs) can be broadly divided into two approaches: one targeting sample or token selection, and the other integrating multiple reward signals to guide generation. DivPO~\cite{lanchantin2025diverse} selects preference pairs of samples online, using `high-quality and rare' generated outputs as positive examples and `common but low-quality' results as negative examples, improving diversity while maintaining generation quality. Cui et al.~\cite{cui2025entropy} addresses entropy collapse by pruning action probabilities and applying KL regularization to high-covariance tokens (action probability and logits changes), helping the policy avoid entropy collapse and enhancing diversity. 
Another line of work integrates multiple reward signals to improve diversity in generation. CPO~\cite{ismayilzada2025creative} modularizes the injection of multiple creative dimensions into the preference optimization objective, adjusting the weight of each dimension to adapt to varying task needs. In the context of image generation, Astolfi et al.~\cite{astolfi2024consistency} use Pareto fronts to compare image generation models. They find that newer models like LDM-Turbo achieve greater consistency and realism but are less diverse, while older models such as LDM offer superior diversity. DiADM~\cite{dombrowski2025image} decouples diversity and quality by introducing a diversity-aware module with pseudo-unconditional feature inputs.
Ding et al.~\cite{ding2023quality} utilize human judgments on similarity and combining latent space projection with contrastive learning to progressively infer diversity metrics. However, it introduces significant complexity due to the multi-stage training process. 
In this work, we addresses the diversity degradation problem from the dual perspectives of distributional reward modeling and generation dynamics. 

\begin{figure*}[t]
  \centering
   \includegraphics[width=\linewidth]{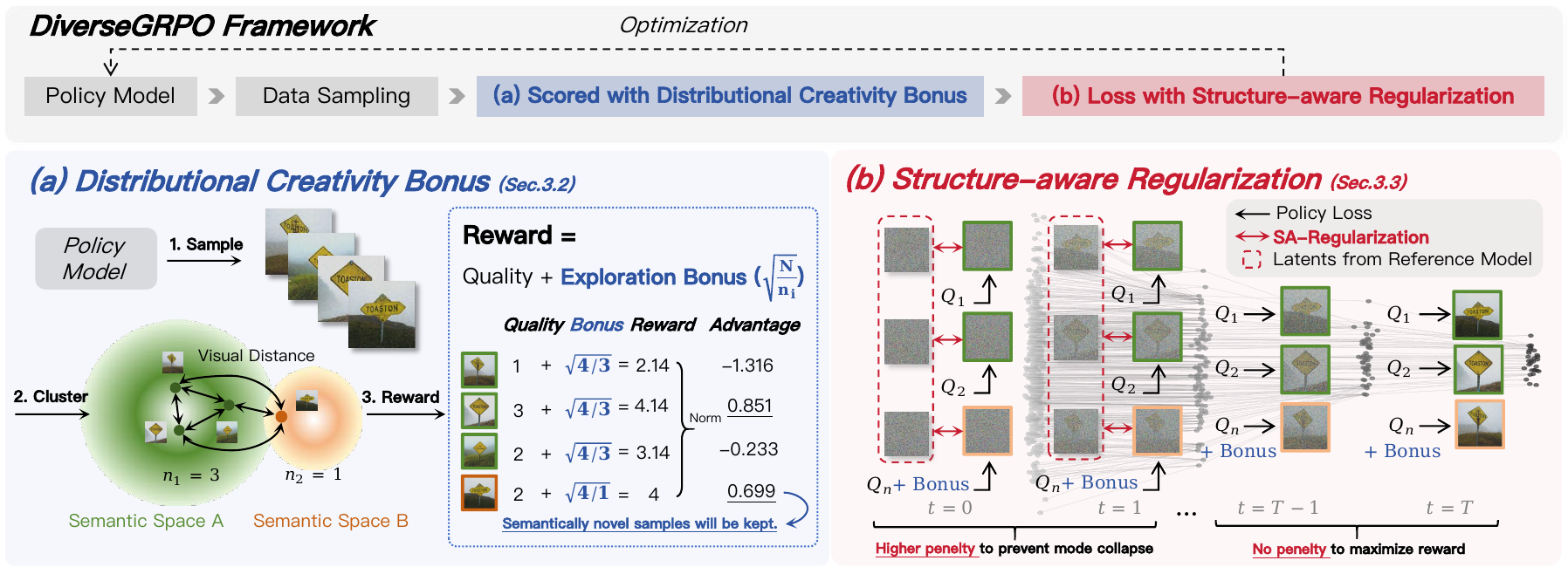}
   \vspace{-15pt} 
   \caption{DiverseGRPO employs two primary strategies to mitigate mode collapse: (a) A distributional creativity bonus mechanism based on semantic grouping. It begins by applying spectral clustering to images generated from the same caption, then assigns exploratory rewards according to cluster size to encourage the emergence of novel visual modes. (b) Structure-aware regularization imposes stronger constraints during the initial denoising stages to preserve sample diversity, while gradually relaxing the penalty in later stages to enhance the effectiveness of reward optimization.}
   \label{DiverseGRPO}
\vspace{-15pt}   
\end{figure*}
\section{Method}
\subsection{Preliminary}
The goal of Reinforcement Learning is to learn a policy that maximizes expected cumulative reward. GRPO achieves this by optimizing its policy model to maximize the following objective:
\begin{equation}
\begin{aligned}
    \mathcal{J}_{\text{Flow-GRPO}}(\theta) = \mathbb{E}_{\boldsymbol{c} \sim \mathcal{C},\,
    \left\{\boldsymbol{x}^{i}\right\}_{i=1}^{G} \sim \pi_{\theta_{\text{old}}}(\cdot \mid \boldsymbol{c})}
    f(r, \hat{A}, \theta, \epsilon, \beta)
\end{aligned}
\end{equation}
\begin{equation}
\begin{aligned}
    f(r, \hat{A}, \theta, \epsilon, \beta) 
    &= \frac{1}{G} \sum_{i=1}^{G} \frac{1}{T} \sum_{t=0}^{T-1} \left( \min \left( r_{t}^{i}(\theta) \hat{A}_{t}^{i}, \right. \right. \\
    &\quad \left. \left. \operatorname{clip}\left( r_{t}^{i}(\theta), 1-\epsilon, 1+\epsilon \right) \hat{A}_{t}^{i} \right) \right) \\
    &\quad \left. - \beta D_{\mathrm{KL}} \left( \pi_{\theta} \| \pi_{\text{ref}} \right) \right), \\
    r_{t}^{i}(\theta) &= \frac{p_{\theta} \left( \boldsymbol{x}_{t-1}^{i} \mid \boldsymbol{x}_{t}^{i}, \boldsymbol{c} \right)}
    {p_{\theta_{\text{old}}} \left( \boldsymbol{x}_{t-1}^{i} \mid \boldsymbol{x}_{t}^{i}, \boldsymbol{c} \right)}, T \text{ is the timestep.}
\end{aligned}
\end{equation}
Given a prompt $\boldsymbol{c}$, the flow model $p_{\theta}$ samples a group of $G$ individual images 
$\left\{ \boldsymbol{x}_{0}^{i} \right\}_{i=1}^{G}$ and the corresponding reverse-time trajectories 
$\left\{ \left( \boldsymbol{x}_{T}^{i}, \boldsymbol{x}_{T-1}^{i}, \ldots, \boldsymbol{x}_{0}^{i} \right) \right\}_{i=1}^{G}$. 
Then, the advantage of the $i$-th image is calculated by normalizing the group-level rewards as follows:
\begin{equation}
\label{advantage_norm}
\begin{aligned}
    \hat{A}_{t}^{i} = \frac{R\left( \boldsymbol{x}_{0}^{i}, \boldsymbol{c} \right) 
    - \operatorname{mean} \left( \left\{ R\left( \boldsymbol{x}_{0}^{i}, \boldsymbol{c} \right) \right\}_{i=1}^{G} \right)}
    {\operatorname{std} \left( \left\{ R\left( \boldsymbol{x}_{0}^{i}, \boldsymbol{c} \right) \right\}_{i=1}^{G} \right)}
\end{aligned}
\end{equation}
Flow-GRPO transforms the original deterministic ODE into an SDE. A key property of this transformation is that the resulting SDE preserves the original model's marginal probability density function at every point in time. The ODE and SDE is as follows:
\begin{equation}
\begin{aligned}
    d\boldsymbol{x}_{t} = \boldsymbol{v}_{t} dt
\end{aligned}
\end{equation}
\begin{equation}
\begin{aligned}
    \boldsymbol{x}_{t+\Delta t} 
    &= \boldsymbol{x}_{t} + \left[ \boldsymbol{v}_{\theta} \left( \boldsymbol{x}_{t}, t \right) 
    + \frac{\sigma_{t}^{2}}{2t} \left( \boldsymbol{x}_{t} + (1-t) \boldsymbol{v}_{\theta} \left( \boldsymbol{x}_{t}, t \right) \right) \right] \Delta t \\
    &\quad + \sigma_{t} \sqrt{\Delta t} \, \boldsymbol{\epsilon}
\end{aligned}
\end{equation}
where $\boldsymbol{\epsilon} \sim \mathcal{N}(0, \boldsymbol{I})$ injects stochasticity and $\sigma_{t} = a \sqrt{\frac{t}{1-t}}$. The KL divergence between $\pi_{\theta}$ and the reference policy $\pi_{\text{ref}}$ is a closed form:
\begin{equation}
\label{kl_loss}
\begin{aligned}
D_{\mathrm{KL}}(\pi_{\theta} \| \pi_{\mathrm{ref}}) 
&= \frac{ \left\| \bar{\boldsymbol{x}}_{t+\Delta t,\theta} - \bar{\boldsymbol{x}}_{t+\Delta t,\mathrm{ref}} \right\|^2 }{ 2\sigma_t^2 \Delta t } \\
&= \frac{ \Delta t }{ 2 } \left( \frac{ \sigma_t(1-t) }{ 2t } + \frac{1}{\sigma_t} \right)^2 \\
&\quad \times \left\| \boldsymbol{v}_{\theta}(\boldsymbol{x}_t, t) - \boldsymbol{v}_{\mathrm{ref}}(\boldsymbol{x}_t, t) \right\|^2
\end{aligned}
\vspace{-5pt} 
\end{equation}
\subsection{Distributional creativity bonus}
After GRPO training, policy models tend to generate a narrow range of outputs that match the surface-level features preferred by the reward model. This limitation stems from reward models' inability to account for distributional diversity. They can only assess the quality of individual images in isolation, failing to recognize the range of valid visual interpretations for a given caption.
To address this, we propose a distributional creativity reward that encourages the model to explore a wider range of visual modes. Our method consists of two stages: (1) distance calculation to quantify the visual differences perceived by humans between generated images and (2) spectral clustering to group the images based on these distances, followed by targeted exploration rewards for the underrepresented clusters.

\textbf{Perception distance calculation.} We begin by defining the pairwise visual distance between images. Given a set of images $ \{x^{1},x^{2},\ldots,x^{n}\} $, wwe compute the perceptual distance between each pair using \textit{DreamSim}~\cite{fu2023dreamsim}, a model designed to align with human visual similarity judgments.

The resulting pairwise distance matrix $D$ is an $ n\times n $ matrix where each element $ D_{ij} $ represents the perceptual distance between images $ x_{i} $ and $ x_{j} $. The diagonal entries satisfy $ D_{ii}=0 $, as the distance between an image and itself is zero. This matrix serves as the basis for subsequent similarity analysis or clustering tasks.

\begin{equation}
D=\begin{pmatrix}
0 & d_{12} & \cdots & d_{1n} \\
d_{21} & 0 & \cdots & d_{2n} \\
\vdots & \vdots & \ddots & \vdots \\
d_{n1} & d_{n2} & \cdots & 0
\end{pmatrix} 
\end{equation}

This matrix D serves as the basis for determining the visual diversity of the generated images.

\textbf{Spectral Clustering.} To effectively partition images based on their perceived differences, we use spectral clustering to divide the images into distinct clusters according to their visual similarity. We begin by constructing an affinity matrix \( A \) using a Gaussian kernel, which measures the similarity between images. The affinity between two images \( x_{i} \) and \( x_{j} \) is defined as:
\vspace{-5pt}
\begin{equation}
A_{ij}=\exp\left(-\frac{d_{i j}^{2}}{2\sigma^{2}}\right)
\vspace{-5pt} 
\end{equation}
where \( \sigma \) is a scaling factor that controls the width of the Gaussian kernel. This affinity matrix nonlinearly maps the pairwise similarities into connection weights, thereby forming a graph that captures the complex intrinsic relationships within the data. Next, we compute the degree matrix \( D \), which is a diagonal matrix where each entry \( D_{ii} \) is the sum of the affinities of image \( x_{i} \) with all other images:
\begin{equation}
D_{ii}=\sum_{j=1}^{n} A_{ij}
\end{equation}
Using the degree matrix \( D \) and the affinity matrix \( A \), we construct the normalized Laplacian matrix \( L \):
\begin{equation}
L=D^{-\frac{1}{2}} A D^{-\frac{1}{2}}
\end{equation}
The Laplacian matrix captures the structure of the graph of images, where nodes (images) that are more similar are strongly connected, and nodes that are less similar are weakly connected. To identify distinct clusters, we perform an eigenvalue decomposition of the Laplacian matrix, obtaining the eigenvectors corresponding to the smallest eigenvalues. These eigenvectors encode the most significant components of the data. The resulting eigenvectors are used to form a new matrix \( V \), where each row corresponds to the eigenvector of an image. The rows of \( V \) are then clustered using k-means clustering to partition the images into \( k \) distinct clusters. This step allows us to identify groups of visually similar images, where each cluster represents a distinct visual mode.

\textbf{Reward Allocation.} After partitioning the images into clusters, we assign exploration rewards based on the cluster sizes. Smaller clusters, which represent underexplored visual modes, receive higher rewards. Specifically, the exploration reward for an image $ x_{i} $ in cluster $ C_{k} $ is inversely proportional to the number of images in that cluster:

\begin{equation}
\label{E}
E_{i}=\sqrt{\frac{N}{n_{k}}}
\end{equation}

where $ n_{k} $ is the number of images in cluster $ C_{k} $, $N$ is the total number of samples with the same caption. Taking the square root further moderates the influence, balancing between rewarding diversity and maintaining model stability. By using Eq.~\ref{E}, we ensure that clusters with fewer images (which are more likely to represent visually distinct modes) receive proportionally higher rewards. This incentivizes the model to generate images that are visually distinct and underrepresented, promoting diversity in the generated outputs.

The final reward for an image $ x^{i} $ is a combination of its quality score $ Q_{i} $ and the exploration reward $ R_{i} $. The two components are weighted by a factor $ \beta $, which balances the emphasis on quality and diversity:

\begin{equation}
\text{R}_{i}=Q_{i}+\beta\cdot E_{i}
\end{equation}

where $ Q_{i} $ represents the intrinsic quality score by reward model, and $ \beta $ controls the influence of the diversity bonus. The combined score guides the model to generate high-quality images that also exhibit greater diversity. We then employ the group-relative advantage (Eq.~\ref{advantage_norm}).
\begin{table*}[t]
\centering
\caption{Comparative evaluation of different backbone models and reward models. 
Higher values ($\uparrow$) indicate better performance for DreamSim, BeyondFID(abbreviated as BFID), ImageReward(abbreviated as ImR), PickScore, and UnifiedReward(abbreviated as UniReward), 
while lower values ($\downarrow$) are better for FID. The Improvement is calculated as $\frac{\text{Ours} - \text{Flow\text{-}GRPO}}{\text{Flow\text{-}GRPO}} \times 100\%$. For the FID metric, where lower values are better, the improvement percentage is calculated as $\frac{\text{Flow\text{-}GRPO} - \text{Ours}}{\text{Flow\text{-}GRPO}} \times 100\%$.}
\label{tab:results}
\renewcommand{\arraystretch}{1.25}
\begin{tabularx}{\textwidth}{@{}lcccccccc@{}}
\toprule
\textbf{Algorithm} & 
\multicolumn{3}{c}{\textbf{Diversity}} & 
\multicolumn{4}{c}{\textbf{Quality}} \\
\cmidrule(lr){2-5} \cmidrule(lr){6-9}
& 
\textbf{DreamSim ($\uparrow$)} & 
\textbf{FID ($\downarrow$)} & 
\textbf{BFID ($\uparrow$)} & 
\textbf{SSIM ($\uparrow$)} & 
\textbf{CLIP ($\uparrow$)} & 
\textbf{ImR ($\uparrow$)} & 
\textbf{PickScore ($\uparrow$)} & 
\textbf{UniReward ($\uparrow$)} \\
\midrule

\rowcolor{gray!5}
\multicolumn{9}{c}{\textbf{SD3.5-M / Pickscore}} \\

Flow-GRPO & 0.1278 & 56.206 & 0.0667 & 0.1701 & 0.3278 & 1.3650 & 0.8809 & 3.5817 \\
\rowcolor{gray!10}
\textbf{Ours} & \textbf{0.1517} & \textbf{43.115} & 
\textbf{0.1895} & \textbf{0.2137} & \textbf{0.3339} & \textbf{1.4009} & 
\textbf{0.8837} & \textbf{3.5852} \\
Improvement & \textcolor{green!60!black}{+18.8\%} & \textcolor{green!60!black}{+23.3\%} &
\textcolor{green!60!black}{+184.2\%} & \textcolor{green!60!black}{+25.6\%} & \textcolor{green!60!black}{+1.9\%} & \textcolor{green!60!black}{+2.7\%} & \textcolor{green!60!black}{+0.3\%} & \textcolor{green!60!black}{+0.1\%} \\

\addlinespace
\rowcolor{gray!5}
\multicolumn{9}{c}{\textbf{Flux.1-dev / Pickscore}} \\

Flow-GRPO & 0.1382 & 68.746 & 0.0766 & 0.1578 & 0.3198 & 1.3497 & 0.8750 & 3.5882 \\
\rowcolor{gray!10}
\textbf{Ours} & \textbf{0.1575} & \textbf{62.505} &
\textbf{0.1059} & \textbf{0.1818} & \textbf{0.3266} & \textbf{1.3959} & 
\textbf{0.8779} & \textbf{3.6039} \\
Improvement & \textcolor{green!60!black}{+13.9\%} & \textcolor{green!60!black}{+9.1\%} &
\textcolor{green!60!black}{+38.2\%} & \textcolor{green!60!black}{+15.2\%} & 
\textcolor{green!60!black}{+2.2\%} & \textcolor{green!60!black}{+3.4\%} & \textcolor{green!60!black}{+0.3\%} & \textcolor{green!60!black}{+0.4\%} \\

\addlinespace
\rowcolor{gray!5}
\multicolumn{9}{c}{\textbf{SD3.5-M / HPSv3}} \\

Flow-GRPO & 0.1625 & 34.040 & 0.0971 & 0.1967 & 0.3309 & 1.3037 & 0.8445 & 3.5825 \\
\rowcolor{gray!10}
\textbf{Ours} & \textbf{0.1851} & \textbf{29.820} &
\textbf{0.1646} & \textbf{0.2103} & \textbf{0.3343} & \textbf{1.3239} & 
\textbf{0.8462} & \textbf{3.5894} \\
Improvement & \textcolor{green!60!black}{+13.9\%} & \textcolor{green!60!black}{+12.4\%} &
\textcolor{green!60!black}{+69.4\%} & \textcolor{green!60!black}{+6.9\%} &
\textcolor{green!60!black}{+1.0\%} & \textcolor{green!60!black}{+1.5\%} & \textcolor{green!60!black}{+0.2\%} & \textcolor{green!60!black}{+0.2\%} \\

\bottomrule
\end{tabularx}
\vspace{-15pt}   
\end{table*}

\subsection{Structure-aware Regularization}
\textbf{Rethinking KL penalty.} To prevent a decline in diversity, existing methods~\cite{liu2025flow,he2025tempflow,li2025branchgrpo} introduce a KL regularization term (as shown in Eq.~\ref{kl_loss}), which constrains the divergence between the Gaussian distributions generated by the current policy model and the original base model at each denoising step. As shown in Fig.~\ref{motivation}, we observe that the diversity of generated samples correlates strongly with the early denoising stages of the diffusion process. Intuitively, these early steps define the global structure and semantic modes of the output distribution, thus constituting a limited diversity budget. Preserving this budget requires strong constraints at early stages, where exploration determines the range of possible generations, while later stages should allow freer adaptation for high-reward refinement. However, in Eq.~\ref{kl_loss}, $\sigma^{2}$ decreases rapidly as denoising progresses, leading to adaptive imbalance: In early stages, large variance $\sigma^{2}$ downscales the KL penalty, weakening regularization when diversity should be preserved. In later stages, small variance amplifies the penalty, excessively constraining high-frequency refinement and discouraging reward-driven improvements. This mismatch violates the desired diversity budget allocation, as it leads to insufficient regularization when global structures form and excessive restriction when only local details are refined.

\textbf{Structure-aware Wasserstein Distance} To address this imbalance, we propose a structure-aware regularization that replaces the KL term with a stage-dependent metric. Specifically, for the first $K$ denoising steps, we apply a Wasserstein Distance constraint between the current and reference policies:
\vspace{-5pt}
\begin{equation}
\begin{aligned}
\mathcal{L}_{\mathrm{WD}}=\frac{ \left\| \bar{\boldsymbol{x}}_{t+\Delta t,\theta} - \bar{\boldsymbol{x}}_{t+\Delta t,\mathrm{ref}} \right\|^2 }{ 2}
\end{aligned}
\end{equation}
which removes the variance denominator in the KL formulation. This modification yields a stronger constraint on early-stage updates, forcing the model to maintain semantic coverage and structural diversity across distinct modes. For later steps($x_{t} > K$), we remove the regularization entirely, allowing the policy to freely optimize toward higher reward fidelity. Formally, the overall regularization is defined as:
\begin{equation}
\mathcal{L}_{\text{reg}}(t)=
\begin{cases}
\frac{ \left\| \bar{\boldsymbol{x}}_{t+\Delta t,\theta} - \bar{\boldsymbol{x}}_{t+\Delta t,\mathrm{ref}} \right\|^2 }{ 2 }, & t \leq K, \\
0, & t > K.
\end{cases}
\end{equation}
\section{Experiments}
\begin{figure*}[t]
  \centering
   \includegraphics[width=\linewidth]{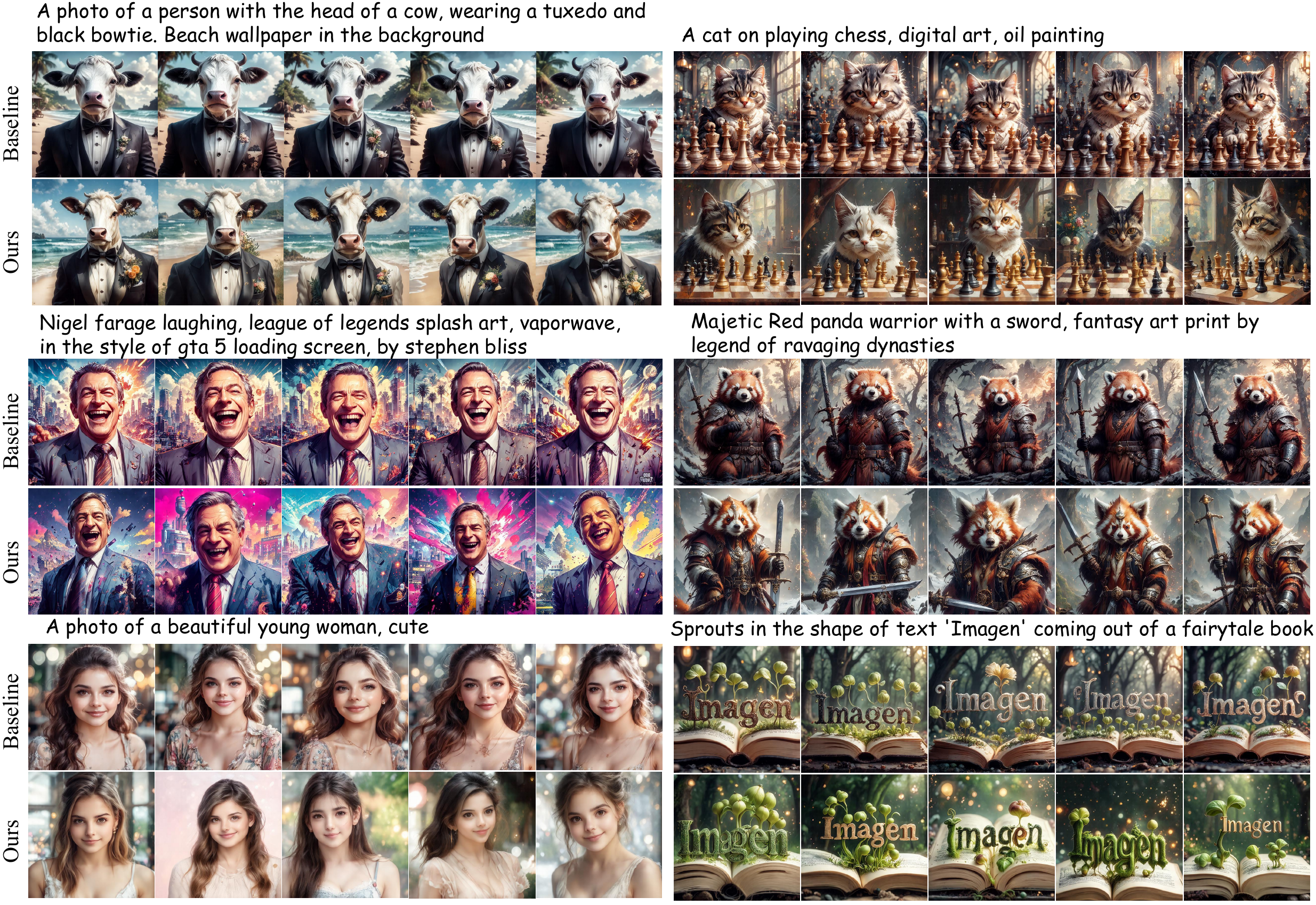}

   \caption{Qualitative experiments on diversity, the baseline method exhibits mode collapse in the generation of main subjects (such as facial features, poses, font colors, etc.), whereas our method achieves greater diversity and creativity while maintaining image quality and consistency with the captions.}
   \label{case}
\vspace{-15pt}   
\end{figure*}
\subsection{Experimental Setting}
\textbf{Implementation Details:} 
We assess the ability of our approach to maintain generation diversity against baseline methods under two model architectures—SD3.5-M~\cite{esser2024scaling} and Flux.1-dev~\cite{blackforestlabs2024flux},and two preference reward functions (Pickscore~\cite{kirstain2023pick} and HPSv3~\cite{ma2025hpsv3}). For the Flux.1-dev architecture, we use 6 steps during training and 28 steps during evaluation, with a classifier-free guidance scale of 3.5. For the SD3.5-M architecture, we use 10 for training and 40 for evaluation, and a guidance scale of 4.5. We apply LoRA fine-tuning to all models, with LoRA rank $r = 32$, scaling factor $\alpha = 64$, learning rate $3\times10^{-4}$, and clip range $1\times10^{-4}$. We use gradient accumulation over $g = 12$ steps and a per-GPU batch size of 2. The exploration reward coefficient $\beta$ is set to 0.7, and regularization coefficient $K$ is set to 4.

\textbf{Evaluation Metrics:} We evaluate the generated images from two perspectives: diversity and quality. 
To quantify sample diversity, we use DreamSim~\cite{fu2023dreamsim}, which measures perceptual similarity between image pairs; lower similarity indicates higher diversity. To further assess the degree of mode collapse, we compute SSIM~\cite{wang2004image}, FID~\cite{heusel2017gans}, and BeyondFID~\cite{dombrowski2025image}. These metrics capture distributional deviations between images generated by the base model and those produced after training. Smaller deviations correspond to weaker collapse and better preservation of the original generative distribution following~\cite{li2025branchgrpo,fu2023dreamsim}.
For image quality, we report CLIPScore~\cite{hessel2021clipscore} to evaluate text–image consistency, together with three human-preference-aligned reward models—PickScore~\cite{kirstain2023pick}, ImageReward~\cite{xu2023imagereward}, and UnifiedReward-Qwen~\cite{wang2025unified}. These metrics collectively measure visual fidelity, semantic alignment, and human-perceived attractiveness.
\begin{figure*}[t]
  \centering
   \includegraphics[width=\linewidth]{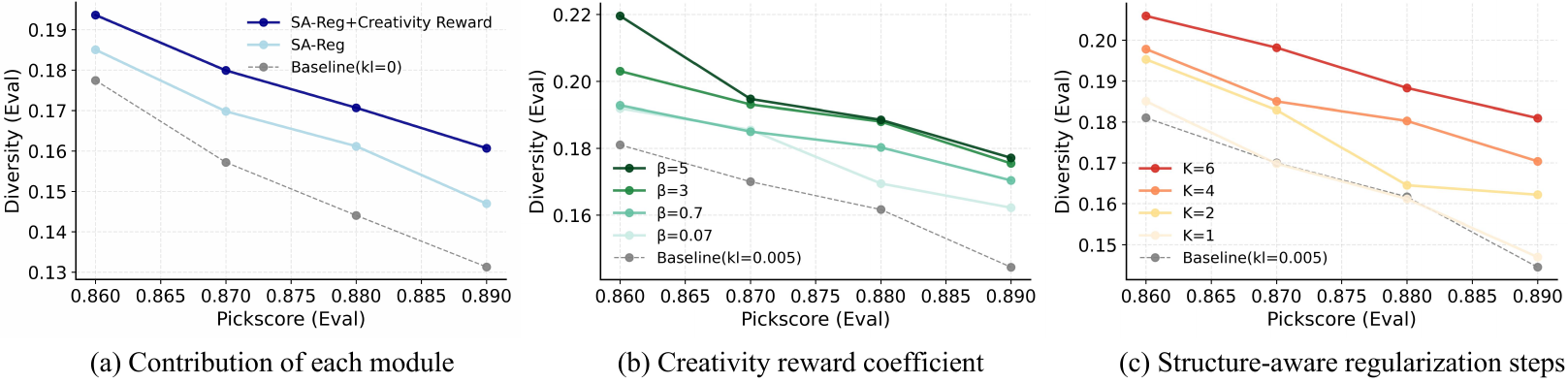}
   \caption{Ablation study on the Pareto front of quality and diversity for different modules and parameters.}
   \label{abla_all}
   \vspace{-10pt}   
\end{figure*}
\begin{figure}[]
  \centering
   \includegraphics[width=\linewidth]{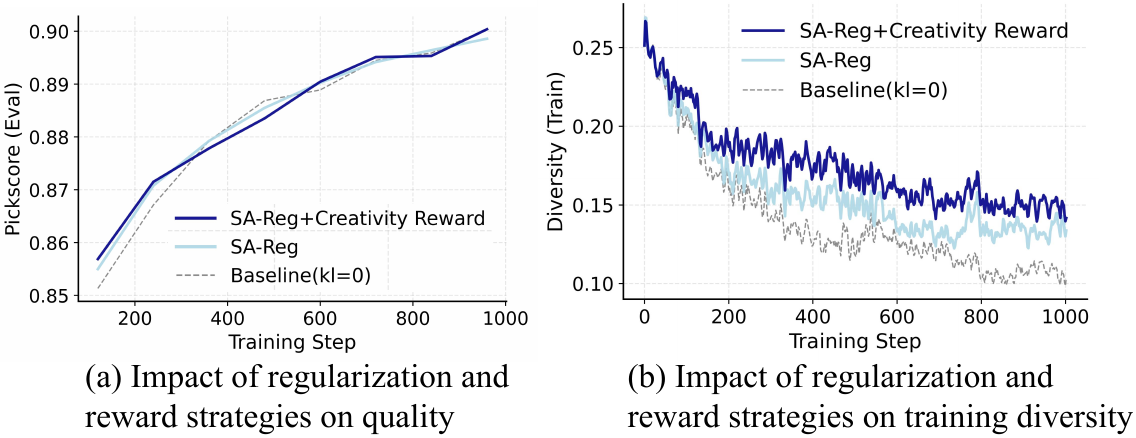}
   \caption{During the training process, DiverseGRPO achieves quality scores comparable to baseline methods, but exhibits a significantly slower decline in diversity.}
   \label{eval_quality}
   \vspace{-10pt}   
\end{figure}
\begin{figure}[t]
  \centering
   \includegraphics[width=\linewidth]{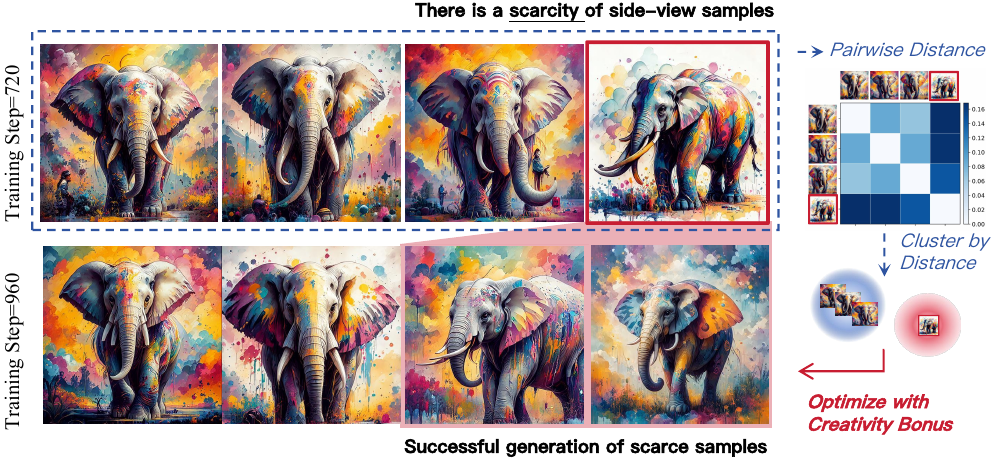}
   \caption{Sample visualization during the training process. Due to the bonus for innovative samples in DiverseGRPO, model can still generate diverse samples in the later stage of training, while it is difficult to mine innovative samples in the baseline (Fig.~\ref{motivation}).}
   \label{vis_process}
   \vspace{-20pt}   
\end{figure}
\subsection{Main Results}
\textbf{Quantitative results:} 
We compare our method's diversity to the baseline Flow-GRPO~\cite{liu2025flow} (The KL term was omitted from the baseline because it significantly slows quality improvement, making a direct comparison of the Pareto fronts infeasible). As shown in Table \ref{tab:results}, our approach consistently enhances all diversity metrics, with improvements of up to +171.4\%in BeyondFID and +18.8\%in DreamSim. This confirms that our reward promotes exploration of novel visual modes, preventing convergence to a few high-reward patterns. The results demonstrate that jointly optimizing for distribution-level exploration and structured generation leads to a more balanced reward optimization process.

\textbf{Qualitative results:} 
As shown in Fig.~\ref{case}, we provide a visual comparison between Flow-GRPO and our method. Under the condition of comparable image quality, our approach demonstrates significantly stronger diversity while remaining faithful to the semantic constraints. The baseline results, on the other hand, exhibit clear signs of mode collapse, generating highly similar and repetitive outputs. For example, given prompts such as `A photo of a beautiful young woman, cute', the baseline model produces images constrained to nearly identical facial expressions and viewpoints. 
In contrast, our method yields diverse stylistic interpretations while still satisfying semantic constraints. 
These visual comparisons conclusively show that our method effectively avoids mode collapse, leading to superior diversity across composition, stylistic details, and dynamic expression.
\subsection{Ablation Study}
\textbf{Contribution of SA-Reg and creativity reward:} We assess the individual contributions of the Structure-Aware Regularization (SA-Reg) and Creativity Reward modules. As shown in Fig.~\ref{abla_all}(a), the combination of both modules (SA-Reg + Creativity Reward) achieves the optimal balance between quality and diversity. Specifically, the SA-Reg module alone enhances diversity, but the addition of the creativity reward significantly boosts the diversity metric further. This suggests that the structure-aware regularization mitigates diversity degradation by encouraging the generation of diverse image patterns, while the creativity reward drives the model to explore even more diverse modes by incentivizing a broader range of semantic groupings. 
Fig.~\ref{eval_quality} shows the image quality and diversity during the training process. Our method exhibits a slower decline in diversity while maintaining image quality comparable to the baseline, demonstrating its effectiveness.

\textbf{Creativity and SA-Reg coefficients:} Figs.~\ref{abla_all}(b) and \ref{abla_all}(c) analyze the impact of the creativity reward coefficient $\beta$ and the number of SA-Reg steps $K$, respectively. Increasing $\beta$ enhances diversity, especially at $\beta$=5, by promoting exploration, but the diversity gain for $\beta$=5 over $\beta$=3 levels off in the later stages. This could indicate that, after a certain point, the model reaches a balance between exploration and exploitation. Similarly, raising $K$ improves diversity through structured-aware regularization, yet higher steps increase computational cost with limited gains (details in appendix).
\subsection{Impact of Exploration Bonus}
Figure~\ref{vis_process} shows the generated samples during the training process. At training step 720, a rare side-view elephant sample appears. Our method uses spectral clustering to identify these rare samples and assigns higher exploration rewards, allowing the model to continue generating such samples in the later stages of training while also improving the quality of the generated outputs. In contrast, the baseline method (as shown in Fig.~\ref{motivation}) faces difficulty in maintaining sample diversity in the later stages of training, even when rare samples are generated. This is due to the fact that the reward model scores individual samples, which fails to recognize the distribution-level creative value, leading to low diversity in the generated samples during later stages of training.
\section{Conclusion}
We revisit the limitations of GRPO-based reinforcement learning in image generation and show that its conventional reward design and regularization scheme inevitably drive the model toward mode collapse in later training stages. Motivated by these, we introduce DiverseGRPO, which integrates a distributional creativity bonus and structure-aware regularization to recalibrate reward and align denoising-stage constraints with diversity preservation. Extensive experiments show DiverseGRPO substantially mitigates mode collapse, and yields a 13\%$\sim$18\% gain in semantic diversity without compromising visual quality, establishing a new Pareto frontier for image generation.
\section{Acknowledgments}
This work was supported in part by the National Natural Science Foundation of China under Grant U24B6012, 62406167, and Kling Team, Kuaishou Technology.
{
    \small
    \bibliographystyle{ieeenat_fullname}
    \bibliography{main}

@article{liu2025flow,
  title={Flow-grpo: Training flow matching models via online rl},
  author={Liu, Jie and Liu, Gongye and Liang, Jiajun and Li, Yangguang and Liu, Jiaheng and Wang, Xintao and Wan, Pengfei and Zhang, Di and Ouyang, Wanli},
  journal={arXiv preprint arXiv:2505.05470},
  year={2025}
}

@article{xue2025dancegrpo,
  title={DanceGRPO: Unleashing GRPO on Visual Generation},
  author={Xue, Zeyue and Wu, Jie and Gao, Yu and Kong, Fangyuan and Zhu, Lingting and Chen, Mengzhao and Liu, Zhiheng and Liu, Wei and Guo, Qiushan and Huang, Weilin and others},
  journal={arXiv preprint arXiv:2505.07818},
  year={2025}
}

@article{bai2022training,
  title={Training a helpful and harmless assistant with reinforcement learning from human feedback},
  author={Bai, Yuntao and Jones, Andy and Ndousse, Kamal and Askell, Amanda and Chen, Anna and DasSarma, Nova and Drain, Dawn and Fort, Stanislav and Ganguli, Deep and Henighan, Tom and others},
  journal={arXiv preprint arXiv:2204.05862},
  year={2022}
}

@article{casper2023open,
  title={Open problems and fundamental limitations of reinforcement learning from human feedback},
  author={Casper, Stephen and Davies, Xander and Shi, Claudia and Gilbert, Thomas Krendl and Scheurer, J{\'e}r{\'e}my and Rando, Javier and Freedman, Rachel and Korbak, Tomasz and Lindner, David and Freire, Pedro and others},
  journal={arXiv preprint arXiv:2307.15217},
  year={2023}
}

@article{black2023training,
  title={Training diffusion models with reinforcement learning},
  author={Black, Kevin and Janner, Michael and Du, Yilun and Kostrikov, Ilya and Levine, Sergey},
  journal={arXiv preprint arXiv:2305.13301},
  year={2023}
}

@article{cui2025entropy,
  title={The entropy mechanism of reinforcement learning for reasoning language models},
  author={Cui, Ganqu and Zhang, Yuchen and Chen, Jiacheng and Yuan, Lifan and Wang, Zhi and Zuo, Yuxin and Li, Haozhan and Fan, Yuchen and Chen, Huayu and Chen, Weize and others},
  journal={arXiv preprint arXiv:2505.22617},
  year={2025}
}

@article{he2025skywork,
  title={Skywork open reasoner 1 technical report},
  author={He, Jujie and Liu, Jiacai and Liu, Chris Yuhao and Yan, Rui and Wang, Chaojie and Cheng, Peng and Zhang, Xiaoyu and Zhang, Fuxiang and Xu, Jiacheng and Shen, Wei and others},
  journal={arXiv preprint arXiv:2505.22312},
  year={2025}
}

@article{yu2025dapo,
  title={Dapo: An open-source llm reinforcement learning system at scale},
  author={Yu, Qiying and Zhang, Zheng and Zhu, Ruofei and Yuan, Yufeng and Zuo, Xiaochen and Yue, Yu and Dai, Weinan and Fan, Tiantian and Liu, Gaohong and Liu, Lingjun and others},
  journal={arXiv preprint arXiv:2503.14476},
  year={2025}
}

@article{fu2023dreamsim,
  title={Dreamsim: Learning new dimensions of human visual similarity using synthetic data},
  author={Fu, Stephanie and Tamir, Netanel and Sundaram, Shobhita and Chai, Lucy and Zhang, Richard and Dekel, Tali and Isola, Phillip},
  journal={arXiv preprint arXiv:2306.09344},
  year={2023}
}

@article{wang2023diffusion,
  title={Diffusion models generate images like painters: an analytical theory of outline first, details later},
  author={Wang, Binxu and Vastola, John J},
  journal={arXiv preprint arXiv:2303.02490},
  year={2023}
}

@article{ho2020denoising,
  title={Denoising diffusion probabilistic models},
  author={Ho, Jonathan and Jain, Ajay and Abbeel, Pieter},
  journal={Advances in neural information processing systems},
  volume={33},
  pages={6840--6851},
  year={2020}
}

@inproceedings{esser2024scaling,
  title={Scaling rectified flow transformers for high-resolution image synthesis},
  author={Esser, Patrick and Kulal, Sumith and Blattmann, Andreas and Entezari, Rahim and M{\"u}ller, Jonas and Saini, Harry and Levi, Yam and Lorenz, Dominik and Sauer, Axel and Boesel, Frederic and others},
  booktitle={Forty-first international conference on machine learning},
  year={2024}
}

@misc{blackforestlabs2024flux,
  author = {{Black Forest Labs}},
  title = {FLUX: Official inference code for FLUX.1 models},
  year = {2024},
  howpublished = {\url{https://github.com/black-forest-labs/flux}},
  note = {Version: commit hash or version number if available}
}

@article{kirstain2023pick,
  title={Pick-a-pic: An open dataset of user preferences for text-to-image generation},
  author={Kirstain, Yuval and Polyak, Adam and Singer, Uriel and Matiana, Shahbuland and Penna, Joe and Levy, Omer},
  journal={Advances in neural information processing systems},
  volume={36},
  pages={36652--36663},
  year={2023}
}

@inproceedings{ma2025hpsv3,
  title={Hpsv3: Towards wide-spectrum human preference score},
  author={Ma, Yuhang and Wu, Xiaoshi and Sun, Keqiang and Li, Hongsheng},
  booktitle={Proceedings of the IEEE/CVF International Conference on Computer Vision},
  pages={15086--15095},
  year={2025}
}

@article{schulman2017proximal,
  title={Proximal policy optimization algorithms},
  author={Schulman, John and Wolski, Filip and Dhariwal, Prafulla and Radford, Alec and Klimov, Oleg},
  journal={arXiv preprint arXiv:1707.06347},
  year={2017}
}

@inproceedings{wallace2024diffusion,
  title={Diffusion model alignment using direct preference optimization},
  author={Wallace, Bram and Dang, Meihua and Rafailov, Rafael and Zhou, Linqi and Lou, Aaron and Purushwalkam, Senthil and Ermon, Stefano and Xiong, Caiming and Joty, Shafiq and Naik, Nikhil},
  booktitle={Proceedings of the IEEE/CVF Conference on Computer Vision and Pattern Recognition},
  pages={8228--8238},
  year={2024}
}

@article{li2025mixgrpo,
  title={Mixgrpo: Unlocking flow-based grpo efficiency with mixed ode-sde},
  author={Li, Junzhe and Cui, Yutao and Huang, Tao and Ma, Yinping and Fan, Chun and Yang, Miles and Zhong, Zhao},
  journal={arXiv preprint arXiv:2507.21802},
  year={2025}
}

@article{wang2025coefficients,
  title={Coefficients-Preserving Sampling for Reinforcement Learning with Flow Matching},
  author={Wang, Feng and Yu, Zihao},
  journal={arXiv preprint arXiv:2509.05952},
  year={2025}
}

@article{he2025tempflow,
  title={Tempflow-grpo: When timing matters for grpo in flow models},
  author={He, Xiaoxuan and Fu, Siming and Zhao, Yuke and Li, Wanli and Yang, Jian and Yin, Dacheng and Rao, Fengyun and Zhang, Bo},
  journal={arXiv preprint arXiv:2508.04324},
  year={2025}
}

@article{li2025branchgrpo,
  title={Branchgrpo: Stable and efficient grpo with structured branching in diffusion models},
  author={Li, Yuming and Wang, Yikai and Zhu, Yuying and Zhao, Zhongyu and Lu, Ming and She, Qi and Zhang, Shanghang},
  journal={arXiv preprint arXiv:2509.06040},
  year={2025}
}

@article{lanchantin2025diverse,
  title={Diverse preference optimization},
  author={Lanchantin, Jack and Chen, Angelica and Dhuliawala, Shehzaad and Yu, Ping and Weston, Jason and Sukhbaatar, Sainbayar and Kulikov, Ilia},
  journal={arXiv preprint arXiv:2501.18101},
  year={2025}
}

@article{ismayilzada2025creative,
  title={Creative preference optimization},
  author={Ismayilzada, Mete and Laverghetta Jr, Antonio and Luchini, Simone A and Patel, Reet and Bosselut, Antoine and van der Plas, Lonneke and Beaty, Roger},
  journal={arXiv preprint arXiv:2505.14442},
  year={2025}
}

@inproceedings{dombrowski2025image,
  title={Image Generation Diversity Issues and How to Tame Them},
  author={Dombrowski, Mischa and Zhang, Weitong and Cechnicka, Sarah and Reynaud, Hadrien and Kainz, Bernhard},
  booktitle={Proceedings of the Computer Vision and Pattern Recognition Conference},
  pages={3029--3039},
  year={2025}
}

@inproceedings{ding2023quality,
  title={Quality diversity through human feedback},
  author={Ding, Li and Zhang, Jenny and Clune, Jeff and Spector, Lee and Lehman, Joel},
  booktitle={Second Agent Learning in Open-Endedness Workshop},
  year={2023}
}

@article{heusel2017gans,
  title={Gans trained by a two time-scale update rule converge to a local nash equilibrium},
  author={Heusel, Martin and Ramsauer, Hubert and Unterthiner, Thomas and Nessler, Bernhard and Hochreiter, Sepp},
  journal={Advances in neural information processing systems},
  volume={30},
  year={2017}
}

@inproceedings{hessel2021clipscore,
  title={Clipscore: A reference-free evaluation metric for image captioning},
  author={Hessel, Jack and Holtzman, Ari and Forbes, Maxwell and Le Bras, Ronan and Choi, Yejin},
  booktitle={Proceedings of the 2021 conference on empirical methods in natural language processing},
  pages={7514--7528},
  year={2021}
}

@article{xu2023imagereward,
  title={Imagereward: Learning and evaluating human preferences for text-to-image generation},
  author={Xu, Jiazheng and Liu, Xiao and Wu, Yuchen and Tong, Yuxuan and Li, Qinkai and Ding, Ming and Tang, Jie and Dong, Yuxiao},
  journal={Advances in Neural Information Processing Systems},
  volume={36},
  pages={15903--15935},
  year={2023}
}

@article{wang2025unified,
  title={Unified reward model for multimodal understanding and generation},
  author={Wang, Yibin and Zang, Yuhang and Li, Hao and Jin, Cheng and Wang, Jiaqi},
  journal={arXiv preprint arXiv:2503.05236},
  year={2025}
}

@article{wang2004image,
  title={Image quality assessment: from error visibility to structural similarity},
  author={Wang, Zhou and Bovik, Alan C and Sheikh, Hamid R and Simoncelli, Eero P},
  journal={IEEE transactions on image processing},
  volume={13},
  number={4},
  pages={600--612},
  year={2004},
  publisher={IEEE}
}

@article{astolfi2024consistency,
  title={Consistency-diversity-realism Pareto fronts of conditional image generative models},
  author={Astolfi, Pietro and Careil, Marlene and Hall, Melissa and Ma{\~n}as, Oscar and Muckley, Matthew and Verbeek, Jakob and Soriano, Adriana Romero and Drozdzal, Michal},
  journal={arXiv preprint arXiv:2406.10429},
  year={2024}
}
}

\end{document}